\documentclass[bdcc,article,accept,pdftex,moreauthors]{Definitions/mdpi} 
\usepackage{subfigure}
\firstpage{1} 
\pubvolume{1}
\issuenum{1}
\articlenumber{0}
\pubyear{2025}
\copyrightyear{2025}
\externaleditor{~} % More than 1 editor, please add `` and '' before the last editor name
\datereceived{24 March 2025 } 
\daterevised{18 April 2025 } % Comment out if no revised date
\dateaccepted{23 April 2025} 
\datepublished{ } 
\hreflink{https://doi.org/} % If needed use \linebreak
\usepackage{chngcntr}
\usepackage{amssymb}
\usepackage{amsmath}
\usepackage{url}
\usepackage{float}

\Title{Transferring Natural Language Datasets Between Languages Using Large Language Models for Modern Decision Support and Sci-Tech Analytical Systems} %EE: Attention: title altered. %AUTHORS: Thank you, OK.

\TitleCitation{Transferring Natural Language Datasets Between Languages Using Large Language Models for Modern Decision Support and Sci-Tech Analytical Systems}

\Author{Dmitrii Popov %MDPI: Please carefully check the accuracy of names and affiliations. %AUTHORS: We checked the accuracy of names and affiliations.
 $^{1,2,3}$\orcidA{}, Egor Terentev $^{1,2}$\orcidB{}, Danil Serenko $^{1,2}$\orcidC{}, Ilya Sochenkov $^{1,3,4}$\orcidD{} and Igor Buyanov $^{1,}$*\orcidE{}}
\AuthorNames{Dmitrii Popov, Egor Terentev, Danil Serenko, Ilya Sochenkov and Igor Buyanov}
\AuthorCitation{Popov, D.; Terentev, E.; Serenko, D.; Sochenkov, I.; Buyanov, I.}
{

}

\address{%
$^{1}$ \quad Federal Research Center ``Computer Science and Control'' of %MDPI: Please check if ``of'' can be a comma, because the address should be  separated by comma from subordinate to superior as required. %AUTHORS: The ``of'' cannot be a comma, as this is an official affiliation.
 the Russian Academy of Sciences \mbox{(FRC CSC RAS)}, Moscow 119333, Russia; popov.dmitriy.p@yandex.ru (D.P.); eterentevd@yandex.ru (E.T.); serenko.d.s@yandex.ru (D.S.); sochenkov@isa.ru (I.S.) %MDPI: We added these email addresses here according to those submitted online at susy.mdpi.com. Please confirm. %AUTHORS: Thank you, we confirm.
\\
$^{2}$ \quad Faculty of Physics and Mathematics and Natural Sciences, RUDN University, %MDPI: The department/school/faculty/campus of this university is required. Please try to provide this information. %AUTHORS: We added the department.
 Moscow 117198, Russia %MDPI: We changed Russian Federation into Russia as required. Please confirm. %AUTHORS: Thank you, we confirm.
\\
$^{3}$ \quad Institute for Information Transmission Problems of %MDPI:  Please check if ``of'' can be a comma, because every parts are separated by comma as required. %AUTHORS: The ``of'' cannot be a comma, as this is an official affiliation.
 the Russian Academy of Sciences (IITP RAS), \mbox{Moscow 127051}, Russia\\
$^{4}$ \quad Ivannikov Institute for System Programming of %MDPI:  Please check if ``of'' can be a comma, because every parts are separated by comma as required. %AUTHORS: The ``of'' cannot be a comma, as this is an official affiliation.
 the Russian Academy of Sciences (ISP RAS), \mbox{Moscow 109004, Russia}
}
\corres{Correspondence: buyanov.igor.o@yandex.ru}

\abstract{The decision-making process to rule R\&D relies on information related to current trends in particular research areas. In this work, we investigated how one can use large language models (LLMs) to transfer the dataset and its annotation from one language to another. This is crucial since sharing knowledge between different languages could boost certain underresourced directions in the target language, saving lots of effort in data annotation or quick prototyping. We experiment with English and Russian pairs, translating the DEFT (Definition Extraction from Texts) corpus. This corpus contains three layers of annotation dedicated to term-definition pair mining, which is a rare annotation type for Russian. The presence of such a dataset is beneficial for the natural language processing methods of trend analysis in science since the terms and definitions are the basic blocks of any scientific field. We provide a pipeline for the annotation transfer using LLMs. In the end, we train the BERT-based models on the translated dataset to establish a~baseline.}


\begin{document}

%%%%%%%%%%%%%%%%%%%%%%%%%%%%%%%%%%%%%%%%%%
% \setcounter{section}{-1} %% Remove this when starting to work on the template.
% \section{How to Use this Template}

% The template details the sections that can be used in a manuscript. Note that the order and names of article sections may differ from the requirements of the journal (e.g., the positioning of the Materials and Methods section). Please check the instructions on the authors' page of the journal to verify the correct order and names. For any questions, please contact the editorial office of the journal or support@mdpi.com. For LaTeX-related questions please contact latex@mdpi.com.%\endnote{This is an endnote.} % To use endnotes, please un-comment \printendnotes below (before References). Only journal Laws uses \footnote.

% The order of the section titles is different for some journals. Please refer to the "Instructions for Authors” on the journal homepage.

\section{Introduction} %MDPI: For software in text,  please state which version of the software was used. For website, please provide the date you accessed the URL in the following format: “URL (accessed on Day Month Year)”. Please check the whole text carefully. %AUTHORS: We added version of the software. We provided the date of access to the all URLs.

In this paper, we develop our approach by extracting definitions of terms being used or introduced. Extracting definitions helps to track the continuity and frequency dynamics of terminology used in scientific and technical documents over time.  In~the absence of large-scale Russian corpora with labeled terms, the~problem is relevant. The~results of this paper can be used as the basis for terminology extraction tools in analytical systems such as iFORA~(\url{https://issek.hse.ru/en/ifora/}, accessed on 24 April 2025), %MDPI: 1. Footnotes are not supported in our journal. We have therefore included this paragraph in the main text. Please confirm. 2. Please provide the date you accessed the URL in the following format: “URL (accessed on Day Month Year)”. %AUTHORS: 1. Thank you, we confirm. 2. We provided the date of access to the URL.
 SciApp~(\url{https://sciapp.ru/}, accessed on 24 April 2025),  Neopoisk~(\url{https://neopoisk.ru/publ/}, accessed on 24 April 2025), and~others. Information plays a crucial role in the decision-making process to rule R\&D in research institutions, universities, and~companies. As~the information overhead grows year by year, the~natural language processing community is challenged to bring a method that allows orientation in this endless amount of papers. To~stay on the cutting edge of the field or quickly get to know the new one, it is nice to have some ways to make it easier. One can say that terms and their definitions are the basic blocks of any study. It would be a handful to have a method that can extract them from literature in order for the learner to get familiar with them. It is also useful when analyzing a large amount of scientific data. To~analyze the development of scientific and technological directions, it is important to identify the terminology used and introduced by the authors. Classical approaches in scientific and technical analytics use methods of extracting terminology from texts~\cite{Lobanova2024IdentifyingAV, Lobanova2023TrendDU, Devyatkin2018MappingTR} in the form of words and~phrases. 

The mainstream approach for classification problems is to use neural networks. Their ability to model complex, nonlinear relationships in data makes them highly effective wide range of applications, including terminology and its definition extraction. Their architecture allows them to learn hierarchical feature representations from raw input, improving performance with increased data and computational resources. Additionally, advances in deep learning, such as attention mechanisms and Transformer architecture, have significantly enhanced their capability to handle text data. The~availability of large datasets and improvements in hardware, particularly GPUs, have further facilitated the training of deep neural networks, making them a preferred choice. Lastly, extensive open-source libraries and community support have accelerated their adoption and implementation across various fields. However, they require a large amount of annotated data to be trained. The~vast number of parameters in neural networks, especially deep learning models, necessitates extensive training data to prevent overfitting and to generalize well to new, unseen data. Labeled data provides the ground truth that helps the network adjust its weights effectively during training, leading to improved accuracy and performance. Additionally, the~diversity present in large datasets aids in capturing the variability in real-world scenarios, ensuring the model's robustness. Insufficient labeled data can lead to poor model performance, making extensive datasets crucial for successful neural network~training. 

One of the recent resources dedicated to the abovementioned task is the DEFT corpus~\cite{Spala2019DEFTAC}. Developing for the SemEval 2020 task~\cite{Spala2020SemEval2020T6} consists of the English texts from free e-books with tree-layer annotation: whether the text has a definition, annotation of terms and definitions as named entities, and~relations between them. % EE: Please check that the intended meaning has been retained. %AUTHORS: confirmed
The~problem is that other languages lack such resources, such as Russian. It would be great to somehow automatically transfer the existing English DEFT dataset into other languages to obtain a starting point. Further, such a transferred dataset could be corrected by human annotators, which is easier and cheaper than crafting the dataset from~scratch.

While general text classification annotation could be transferred by the language translation, the~transferring of the named entity annotation from one language to another is challenging since the matched spans in the source and target languages must be found. Recently, the~large language models (LLMs) have demonstrated high effectiveness in a broad variety of NLP tasks. Unlike traditional NLP methods that rely heavily on manual feature engineering and rule-based systems, LLMs leverage the Transformer architecture to~learn patterns and contextual relationships from vast amounts of text data. This approach allows LLMs to perform tasks such as translation~\cite{Zhu2023MultilingualMT} and named entity recognition (NER) \cite{Zhou2023UniversalNERTD} with greater efficiency and adaptability across different languages and contexts. LLMs surpass traditional methods by eliminating the need for extensive task-specific programming and by excelling in zero-shot and few-shot learning scenarios, where little to no task-specific data are available. They bring improvements in scalability, as~they can be fine-tuned for numerous applications with minimal adjustments, and~they enhance performance by capturing nuanced language subtleties that traditional models may overlook. Furthermore, LLMs have shown the ability to generalize across tasks, providing a unified model capable of addressing diverse NLP challenges. This versatility reduces the need for multiple specialized systems, ultimately streamlining the development process. Motivated by these advantages, we utilize LLMs for the cross-language annotation transfer by using them as a ``smart'' translator that can preserve the named entity spans while translating the text. We hypothesize that they can automate the process of adaptation of an annotated dataset from different languages, allowing one to obtain a quick baseline or giving a solid start in the annotation process. We limit this work only to English--Russian language pairs, leaving other languages for future~work.

To summarize, the~contribution of our work is as~follows:
\begin{itemize}
    \item We show how we transfer the NER annotation using LLMs on the English and Russian language pairs.
    \item We analyze the translation quality of several modern LLMs from English to Russian for this particular task. It includes ChatGPT, Llama3.1-8B~\cite{Dubey2024TheL3}, DeepSeek, and~Qwen. 
    \item We provide the result of the BERT-like models trained to make a baseline for two of the three original DEFT tasks: detection of texts with definitions and named entity recognition. In~addition, we provide the results of our pipeline for the Wikipedia part of the WCL dataset.
\end{itemize}

We opensourced the datasets~(\url{https://huggingface.co/datasets/astromis/ruDEFT}, accessed on 24 April 2025,  \url{https://huggingface.co/datasets/astromis/WCL_Wiki_Ru}, accessed on 24 April 2025) and code~(\url{https://github.com/Astromis/research/tree/master/rudeft}, accessed on 24 April 2025). 

\section{Related~Work}
The task of term and definition extraction has a long story because it is tightly related to the desire to structure the information from various texts. Starting from rule-based systems~\cite{Klavans2001EvaluationOT}, which relies on handcrafted rules, it evolves to statistical methods~\cite{Navigli2010LearningWL, Frantzi1999TheCD,Sun2023DiscoveringPO} and lastly to a deep learning system~\cite{Collard2023ParmesanMC,Kang2020DocumentLevelDD}. The~statistical methods operate either by automatically mining the patterns from the dataset or by constructing a set of features that fit into a machine learning model. The~deep learning methods fully rely on neural networks like LSTM~\cite{Hochreiter1997LongSM} or, recently, Transformer-based architecture~\cite{Vaswani2017AttentionIA}.

Frequently, statistical or deep learning-based papers come with their datasets. The~WCL dataset~\cite{Navigli2010AnAD} was developed as a part of the work in~\cite{Navigli2010LearningWL}. This is a dataset with annotated definitions and hypernyms composed of Wikipedia pages and a subset of the ukWaC Web corpus. The~SymDef dataset comes from~\cite{MartinBoyle2023ComplexMS} and approaches the problem of bounding the mathematical symbols with their definitions in scientific papers parsed from arXiv. Speaking about the lack of resources, it is worth mentioning that for the Russian language, it is easy to find only the RuSERRC dataset~\cite{Bruches2020EntityRA} in which the terms were annotated. The~Russian datasets with definition annotation are~unknown.

With the rise of the LLMs, researchers have started to investigate the method of using them in dataset annotation or generation. In~the work~\cite{He2023AnnoLLMML} researchers employ ChatGPT to be an annotator with an explain-than-annotate technique. They compared its performance with crowdsourcing annotation and obtained promising results that ChatGPT is on par with crowdsourcing. Regularly, researchers issue the best practices about how to obtain the best from  LLMs as annotators, like in~\cite{Alizadeh2023OpenSourceLF}. LLMs are used in interesting annotation projects, like creativity dataset creation~\cite{Tian2023MacGyverAL}, where the LLM produces some ideas of how to use an unrelated set of items to solve a particular task, and humans only verify these ideas. In~some cases, the~LLM is used directly for dataset synthesis~\cite{Ghanadian2024SociallyAS, Frei2023AnnotatedDC}.

%%%%%%%%%%%%%%%%%%%%%%%%%%%%%%%%%%%%%%%%%%
\section{Materials and~Methods}
\unskip

\subsection{DEFT~Corpus}

\begin{figure}[H]

\includegraphics[width=\textwidth,height=\textheight,keepaspectratio]{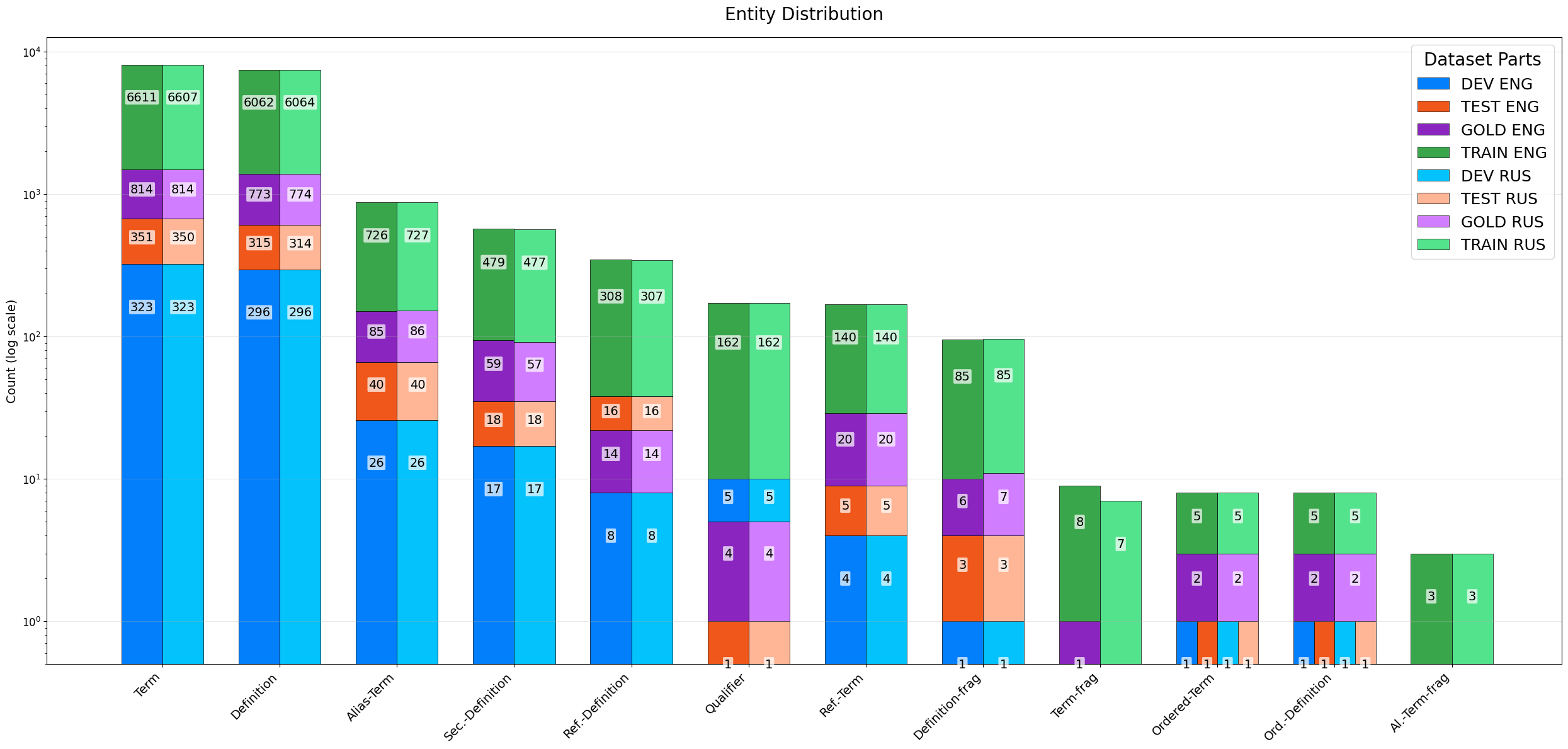}
\caption{The statistics of the entities in the DEFT corpus. The gold part shows the part prepared by us for testing the~LLMs.}\label{deft_stat}
\end{figure}

The DEFT corpus is a collection of text from free books available on %MDPI: Please provide the date you accessed the URL in the following format: “URL (accessed on Day Month Year)”. %AUTHORS: We provided the date of access to the URL.
 \url{https://cnx.org/} (accessed on 24 April 2025). The~texts cover topics like biology, history, physics, psychology, economics, sociology, and~government. NER annotation includes the term and definition labels as primary annotation and supportive annotation for the cases as aliases, orders, and~referents for both terms and definitions. % EE: Please check that the intended meaning has been retained. %AUTHORS: confirmed
 For~the task of detecting a sentence that contains a definition, the~annotation is produced straightforwardly from the presence of the definition NER annotation in a sentence. The~relation annotation bounds terms and definitions. For~a detailed description of the tags, annotation process, and~challenges, we refer to the original paper. We use the available DEFT corpus on GitHub~(\url{https://github.com/adobe-research/deft_corpus}, accessed on 24 April 2025, version of the corpus from 16 January 2020, commit is db8c95565c2e58d861537cb8cb4621c50b75cd13). %MDPI: 1. Footnotes are not supported in our journal. We have therefore included this paragraph in the main text. Please confirm. 2. Please provide the date you accessed the URL in the following format: “URL (accessed on Day Month Year)”. 3. Please state which version of the software was used. %AUTHORS: 1. Thank you, we confirm. 2. We provided the date of access to the URL. 3. We added version of the corpus and commit.
 The~entity statistics are provided in Figure~\ref{deft_stat} in "ENG" legend parts for train, dev test, and~the prepared gold set, which we talk about in the section ``Preparing the gold set''. We also want to point out that we will not experiment with a third annotation with a relation between terms and definitions. We leave it for future~work.

% % NEW FORMAT
% \begin{table}[H]
% \centering
% \caption{The statistics of the entities in DEFT corpus. GOLD part shows the part prepared by ourselves for testing the LLMs.}
% \label{deft_stat}
% \begin{tabularx}{\textwidth}{@{}l*{9}{C}@{}}
% \toprule
% \textbf{Entity name} & \multicolumn{2}{c}{\textbf{DEV}} & \multicolumn{2}{c}{\textbf{TRAIN}} & \multicolumn{2}{c}{\textbf{GOLD}} & \multicolumn{2}{c@{}}{\textbf{TEST}} \\
% \cmidrule(lr{0.5em}){2-3} \cmidrule(lr{0.5em}){4-5} \cmidrule(lr{0.5em}){6-7} \cmidrule(lr{0.5em}){8-9}
%  & \textbf{ENG} & \textbf{RUS} & \textbf{ENG} & \textbf{RUS} & \textbf{ENG} & \textbf{RUS} & \textbf{ENG} & \textbf{RUS} \\
% \midrule
% Term            & 323 & 323 & 6611 & 6607 & 814 & 814 & 351 & 350 \\
% Definition      & 296 & 296 & 6062 & 6064 & 773 & 774 & 315 & 314 \\
% Alias-Term      & 26  & 26  & 726  & 727  & 85  & 86  & 40  & 40  \\
% Sec.-Definition & 17  & 17  & 479  & 477  & 59  & 57  & 18  & 18  \\
% Ref.-Definition & 8   & 8   & 308  & 307  & 14  & 14  & 16  & 16  \\
% Qualifier       & 5   & 5   & 162  & 162  & 4   & 4   & 1   & 1   \\
% Ref.-Term       & 4   & 4   & 140  & 140  & 20  & 20  & 5   & 5   \\
% Definition-frag & 1   & 1   & 85   & 85   & 6   & 7   & 3   & 3   \\
% Term-frag       & 0   & 0   & 8    & 7    & 1   & 0   & 0   & 0   \\
% Ordered-Term    & 1   & 1   & 5    & 5    & 2   & 2   & 1   & 1   \\
% Ord.-Definition & 1   & 1   & 5    & 5    & 2   & 2   & 1   & 1   \\
% Al.-Term-frag   & 0   & 0   & 3    & 3    & 0   & 0   & 0   & 0   \\
% \bottomrule
% \end{tabularx}
% \end{table}

To work with our pipeline, we had to convert the original CoNLL-like (Conference on Natural Language Learning) format into the Hugging Face Datasets library~\cite{Lhoest2021DatasetsAC}. Basically, the~dataset can be represented as a list of dictionaries (objects) that can be easily converted to and from JSON. In~turn, it makes it easy to communicate with LLMs in the latter~format. 

We noticed two issues while preprocessing the original files of the corpus. The~first one is a tokenization error when two sentences containing the wide span are wrongly separated. As~a consequence, the~second sentence starts with the token having an ``I'' tag, which is illegal in the IOB (inside, outside, beginning) format.

While the first issue was found just once, the~second issue with data duplicates occurs more often. We count 2187 duplicate sentences. Moreover, these duplicates have different annotations. It seems that the merging error occurred when the corpus was being~compiled.

\subsection{WCL~Corpus}

In addition, to make our research broader, we apply our pipeline to the Wikipedia part of the WCL dataset. We chose this part because of a clear understanding of the structure, where all data were divided into two files: one file contains sentences with definitions and another file contains just regular sentences. All these sentences have an annotation of the term token, but not for the definition~itself.

We convert the dataset from its original format to the common Hugging Face structure, resampling what we obtain from converting the DEFT dataset. All in all, we obtain 2822 sentences with no definitions and 1869 sentences with definitions. We next divide the whole dataset into train, dev, and test splits in the proportion 70/10/20.

Nevertheless, our main focus is the DEFT~dataset.

\subsection{Large Language~Models}
\label{subsec:llms}

We benchmarked a diverse set of state‑of‑the‑art LLMs to assess their performance on our tasks. The~specific model checkpoints evaluated~are as follows:
\begin{itemize}
  \item llama-3.1-8b-instruct %MDPI: it's suggested to unify all the font to be Palatino Linotype. We highlighted only a few cases as a reminder, please check the whole text carefully. %AUTHORS: We replaced all model names to the font Palatino Linotype.

  \item gpt-3.5-turbo
  \item gpt-4o-mini
  \item gpt-4.1-nano
  \item gpt-4.1-mini
  \item deepseek-chat-v3-0324
  \item deepseek-r1
  \item qwen-2.5-72b-instruct
\end{itemize}

To simplify the experimental setup and ensure reproducibility, we leveraged the bothub.chat service (\url{https://bothub.chat/}, accessed on 24 April 2025) %MDPI: Please provide the date you accessed the URL in the following format: “URL (accessed on Day Month Year)”. %AUTHORS: We provided the date of access to the URL. 
 as a unified proxy for accessing the aforementioned models. This service provides a streamlined interface to various APIs---including OpenAI’s ChatGPT, DeepSeek, Llama, and~QWEN---thereby abstracting the need for direct API integration. This approach not only facilitated rapid testing and experimentation but also allowed for systematic documentation of any associated computational costs, which were primarily linked to underlying API usage fees. Detailed statistics on time and monetary expenditures are provided in Appendix~\ref{appendix:costs}.

\subsection{Methodology}
\begin{figure}[H]
\includegraphics[width=\textwidth,height=\textheight,keepaspectratio]{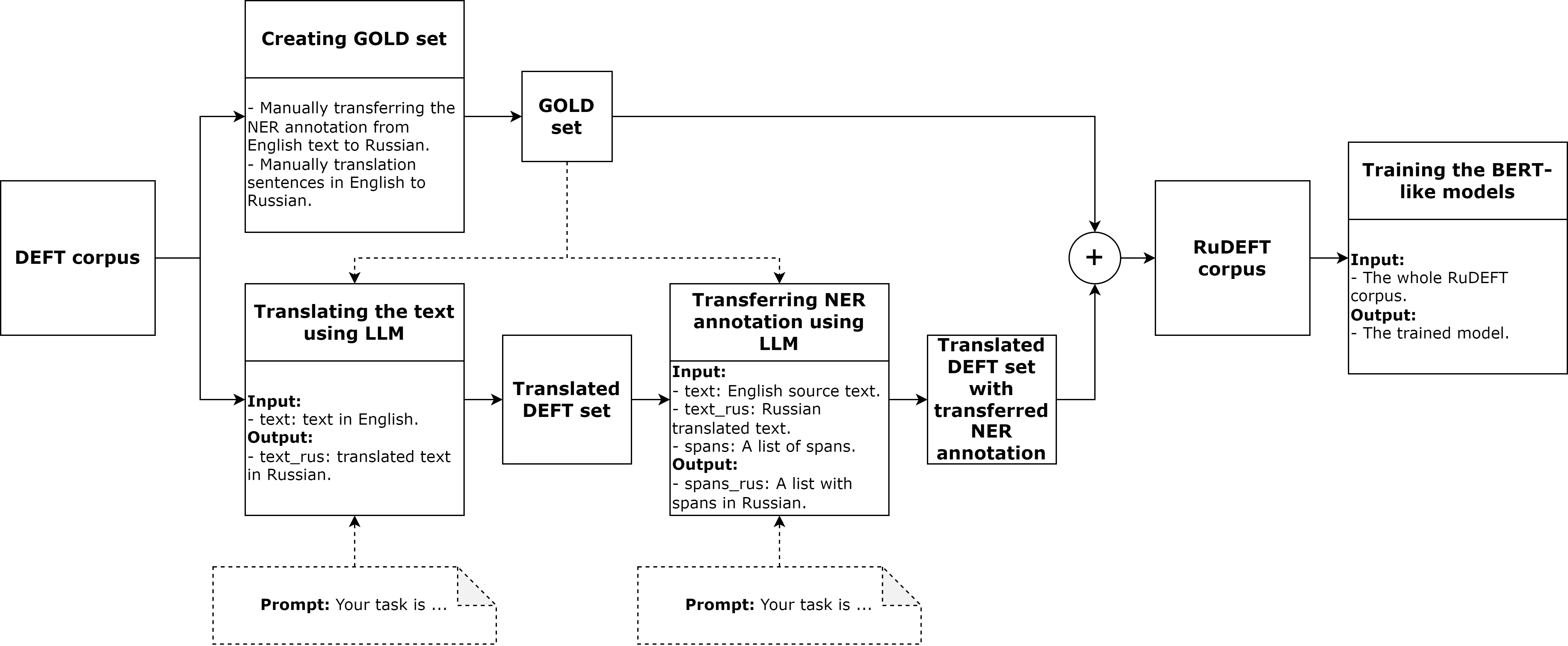}
\caption{The %MDPI: The contents of this figure are not legible. Please replace the image with one of a sufficiently high resolution (min. 1000 pixels width/height, or a resolution of 300 dpi or higher). %AUTHORS: We replaced the image with one of a sufficiently high resolution (400 dpi).
 methodology~steps.}\label{methd_overview}
\end{figure}

On a high level, the~methodology consists of three steps: preparing the gold set, automatic translation, and~annotation transfer. We describe each of them in separate paragraphs. The~overview of the methodology steps is visualized in Figure~\ref{methd_overview}. 

\unskip

\subsubsection{Preparing the Gold~Set}

To be able to estimate the output quality of our steps, we need a reliable set that was manually checked in terms of the translation and NER annotation. To~do that, we translate the whole dev set and a small part of the train set with the API Google Translate. We obtain 1179 sentences from the dev set and 3010 out of 24,184 %MDPI: We added commas to separate out the thousands for numbers with five or more digits. Please confirm. %AUTHORS: Thank you, we confirm.
 randomly selected sentences from the train set. (There is no specific reason why exactly 3010 from the train part were sampled; it just happened once, and~we decided to let it be.) Next, we select only the sentences with NER annotation, which gives us 870 sentences from the dev and train sets. Then we manually transfer the NER annotation from English text to Russian using Label Studio~\cite{Label_Studio}. While transferring, when we saw that the translation was semantically incorrect, we skipped these sentences and later translated them manually. The~source of incorrectness originates commonly from the catchphrases and the specific language. The~statistics of the NER labels of the gold set are available in Figure~\ref{deft_stat}.

For the purpose of evaluating the translation, we sampled 200 sentences with no NER annotation and also checked them for adequate~translation.

Finally, the~statistics of our gold set are next: 870 sentences have the NER annotation, and~200 sentences do not. Overall, we have 1070 sentences with manually verified translations into~Russian.

\subsubsection{Transferring NER Annotation Using~LLMs}

For the task of NER annotation transferring, we test several LLMs, such as Llama3.1-8B and Qwen-2.5-72B, variants of DeepSeek, and variants of GPT-4 and ChatGPT3.5-turbo from OpenAI. They vary in scale, which directly influences the cost and generation time. The~latter is crucial when one works with a large amount of data. Also, while OpenAI's models are closed-sourced, it is of high interest how the open-sourced models such as Llama3.1, Qwen, or DeepSeek are suitable for such tasks, as~many researchers and companies cannot rely on third-party API because of data~privacy.

As a note, we use Qwen2.5-72B because it was released during this work, so we decided to include a bigger LLM of the fresh release and not~Llama3.1-72B.

To build the prompt, we try several standard prompt techniques like zero-shot, few-shot, and~chain-of-thoughts~\cite{Wei2022ChainOT}. We select 20 examples from the gold dataset in the early stages of the prompt development. That means we reject some prompt-building strategies if they cannot deal with most of this subset. The~only prompts that achieve more than 15~cases of correct annotation transferring will be selected for further~testing.

\begin{figure}[H]

\includegraphics[width=\textwidth,height=\textheight,keepaspectratio]{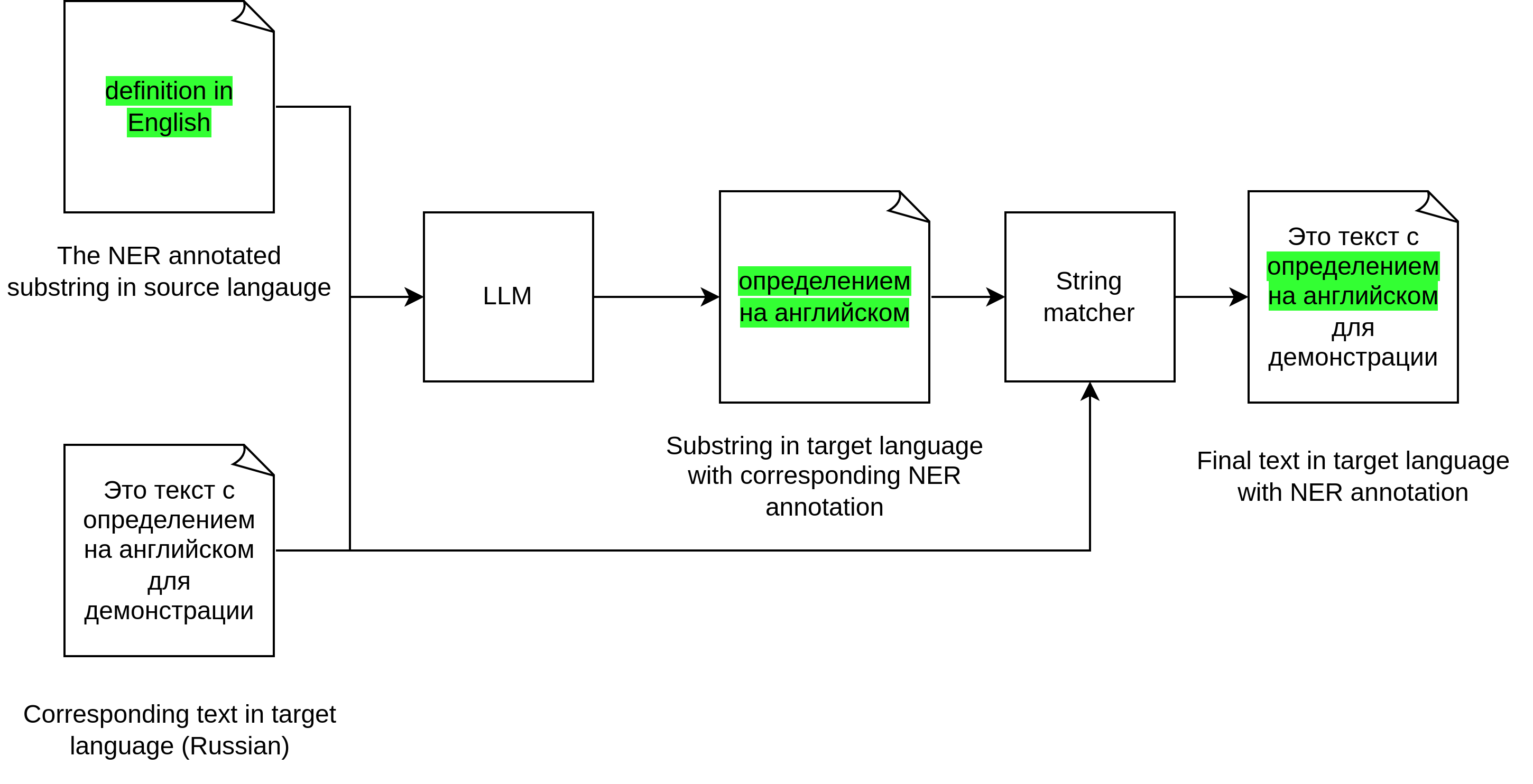}
\caption{The %MDPI: Please remove the non-English term from the figure or add a definition for it. %AUTHORS: We added the definition for the non-English terms, but Russian characters are not supported by your template and we have commented out the line with a note. If the editors could add Russian characters here, then please uncomment the our note.
 step-by-step illustration of the NER annotation transferring. The~green highlight shows the NER~annotation.
    % \footnotesize\textit{Note}: The Russian phrases ``Это текст с определением на английском для демонстрации`` (translated as ``this is a text with a definition in English'') and ``определение на английском`` (``definition in English'') are included for illustrative purposes.  
 }\label{ner_transferring}
\end{figure}

At first, we formulate the task for the LLM as follows: given the source English text and the list of annotation span triplets consisting of start index, end index, and~label name, we ask the model to find in a given Russian text span triplets that correspond to English ones. Unfortunately, this approach failed to give a good output quality, as~we  show in the ``Results'' section. %MDPI: There is no ``Results'' section'' of this paper. Please check. %AUTHORS: We checked, and this paper has a "Results" section with number 4.
 So, we reformulated the task to find a substring in Russian that corresponds to a substring in English that is actually an NER span. See Figure~\ref{ner_transferring} for a visual explanation of the resulting~approach.

After obtaining a prompt that beats the simple test, we evaluate it on the whole gold dataset, where we manually transfer the annotation. As~a metric, we use the number of matches between the gold transfer and LLM in different~situations:
\begin{itemize}
    \item Exact matches count the cases when the indices of the gold and transferred spans are~equal;
    \item Wider partial matches count the cases when the transferred span is wider than the original one;
    \item Narrower partial matches count the cases when the transferred span is narrower than the original one;
    \item Mismatches are self-explained cases;
    \item Total spans checked accounts for the processed cases. Note that it differs between LLMs, as some examples could not be handled correctly, even after several retries.
\end{itemize}

\subsubsection{Translating the Text Using~LLMs}
The translation task is pretty straightforward. Given the text in English, we ask the LLM to translate it into Russian. From~the previous research~\cite{Zhu2023MultilingualMT}, we know that LLMs are good enough in this task, though~they still perform worse than supervised systems. Nevertheless, we are interested in building a monolithic pipeline based solely on LLMs. To~ensure the quality of the translation, based on our gold set of translated DEFT, we test the translation abilities of our chosen LLMs with a BLEU score and two metrics based on~embeddings.

The BLEU score~\cite{Papineni2002BleuAM} is a widely used metric in machine translation. The~mechanism of this metric is to calculate the overlap between n-grams of the gold translation and the translation provided by the system. The~known disadvantage of this approach is that lexical overlap does not guarantee meaning preservation. To~estimate to which the sense is preserved, we use embedding-based metrics as they operate on a semantic~level. 

We use the LaBSE model~\cite{Feng2020LanguageagnosticBS} as a cross-lingual encoder for texts, as~it encodes semantically close texts in different languages to close points in one embedding space. This allows us to measure the translation quality in two~ways. 

First, we calculate the mean distance between the corresponding English text and its gold Russian translation, then we do it in the same way between the English text and the translated one. Next, we compare two means by substituting the latter for the former. The~closer to the zero metric is, the~more likely that model-translated text conveys the same meaning as the gold-translated text. We call this metric Parallel Comparison. %MDPI: Please confirm if the bold formatting is necessary; if not, please remove it. The following highlights are the same. %AUTHORS: We removed the bold formatting.

Second, resembling the BLEU approach, we compare the mean distances between gold Russian text embeddings and translated ones. If~the BLEU operates on a lexical level, this metric does this on a semantic level and relies on the abovementioned property of cross-lingual encoders. We name this metric  BLEU-like.

We reuse the best prompt from the annotation transferring task, only changing the task description. As~we will show in the result section, they perform similarly, so we provided the one that we use in Appendix \ref{app_translation_prompt_2}.

%%%%%%%%%%%%%%%%%%%%%%%%%%%%%%%%%%%%%%%%%%
\section{Results}
\unskip

\subsection{Annotation~Transferring}
As mentioned earlier, we try to transfer NER annotation by asking LLMs to write triplets in Russian tasks according to the original English spans. However, the~LLMs failed to provide a good result in every setting of prompts and even when using ChatGPT-4. The~best result we managed to achieve is only 2 out of 20 testing subset examples with this model. We notice several issues related to this~failure:
\begin{itemize}
    \item Wrong index determination: The model did not use the provided indices for exact text extraction. It tried to figure out by itself which text should be extracted instead.
    \item Span length mismatch: When the model tried to follow the provided indices, the~extracted text length did not match with actual span text.
    \item Ineffectiveness of the task correction: The efforts to explain the task more precisely did not bring valuable improvements.
    \item Failures of the self-generated prompts: We try to generate prompts by the model itself after providing the detailed task description. However, it also did not help the model to get better at using indices.
\end{itemize}

After changing the task formulation to extract a substring directly, in~a short time, we found a 2-shot prompt that can correctly solve 18 out of 20 testing tasks with ChatGPT3.5-turbo and Llama3.1-8B (we did not test other LLMs, as they were added in the late stage of the work). We chose to use this prompt in the next experiments. The~text of the prompt can be found in Appendix \ref{app_ner_transferring_prompt}. 

While testing the Llama on the gold set, we noticed two~things:
\begin{itemize}
    \item The model tends to break the data format or even send the code, as~we would ask her to write a function to transfer the spans by the available index. Generally, this can be fixed just by resending the request.
    \item The model makes way more unmatchable mistakes. Analysis shows that the source of them is a rich Russian morphology. Sometimes the model corrects the mistakes of the Russian text that was automatically translated. To~address this issue, we apply fuzzy matching as a post-process for Llama3.1.
\end{itemize}

The metric results on the whole gold set for this prompt and the models are provided in Table~\ref{ner_transferring_result}. As~seen, the~deepseek-chat-v3-0324+fs model shows the best results, where only 0.84\% of completely mismatched spans are in common. The~comparable results show other LLMs except for gpt-4.1-nano and llama-3.1-8b-instruct. As~for the latter, we see that it has a much bigger percentage of mismatched spans than the other LLMs and only a little more than half of exact matches. % EE: Please check that the intended meaning has been retained. %AUTHORS: confirmed
However, the~post-processing fuzzy search can effectively reduce the number of mismatches at the cost of a nonexact mismatch rising. % EE: Please check that the intended meaning has been retained. %AUTHORS: confirmed
We hypothesize that fine-tuning the post-processing could further improve the result. Generally, we think that if one makes the instruction tuning of the Llama, it could show a much stronger result. It can certainly be found in cases where this strategy makes sense, considering the much lower inference cost of small~models.

% NEW FORMAT
% \begin{table}[H]
% \centering
% \caption{The results of the NER transferring. We omit some model details to narrow down the table. fs means "fuzzy search".}
% \label{ner_transferring_result}
% \begin{tabularx}{\textwidth}{@{}l*{6}{C}@{}}
% \toprule
% \textbf{} & \multicolumn{3}{c}{\textbf{Dev}} & \multicolumn{3}{c@{}}{\textbf{Train}} \\
% \cmidrule(lr){2-4} \cmidrule(l){5-7}
% \textbf{} & GPT-3.5 & Llama & Llama+fs & GPT-3.5 & Llama & Llama+fs \\
% \midrule
% Total Entries            & 239  & 239  & 239   & 526  & 526  & 526  \\
% Exact Match        & 454  & 368  & 398   & 1019 & 768  & 836  \\
% Wider Match    & 23   & 19   & 28    & 46   & 54   & 76   \\
% Narr. Match     & 10   & 19   & 33    & 20   & 71   & 100  \\
% Mismatched         & 8    & 83   & 30    & 36   & 182  & 63   \\
% Spans Checked      & 494  & 489  & 489   & 1118 & 1068 & 1068 \\
% \bottomrule
% \end{tabularx}
% \end{table}

% new table with %

\begin{table}[H]\small
\centering
\caption{The %MDPI: 1. Table should be cited close to where it is first mentioned. We moved the table here. Please confirm. 2. We changed the commas between digits into decimal dots. Please confirm these revisions. 3.  Please add an explanation for the use of bold in the table footer. If the bold is unnecessary, please remove it. %AUTHORS: 1. Thank you, we confirm. 2. Thank you, we confirm. 3. We added the explanation about using bold values in the table.
 results of the NER transferring. We omit some model details to narrow down the table. fs means ``fuzzy~search''.}
\label{ner_transferring_result}
\begin{tabularx}{\textwidth}{@{}l CCCCC @{}}
\toprule
\textbf{Model} & \textbf{Exact Match (\%)} & \textbf{Wider Match (\%)} & \textbf{Narrower Match (\%)} & \textbf{Mismatched (\%)} & \textbf{Spans Checked (\%)} \\
\midrule
\textbf{llama-3.1-8b-instruct} & 66.57 & 7.02 & 4.72 & 12.42 & 90.45 \\
\textbf{llama-3.1-8b-instruct+fs} & 71.18 & 8.31 & 7.08 & 4.16 & 90.45 \\
\textbf{gpt-3.5-turbo} & 90.62 & 4.27 & 1.97 & 3.09 & 99.72 \\
\textbf{gpt-3.5-turbo+fs} & 91.63 & 4.78 & 2.08 & 1.46 & 99.72 \\
\textbf{gpt-4o-mini} & 86.91 & 10.00 & 0.96 & 2.13 & \textbf{99.78} \\
\textbf{gpt-4o-mini+fs} & 87.42 & 10.17 & 1.12 & 1.29 & \textbf{99.78} \\
\textbf{gpt-4.1-nano} & 69.72 & 9.44 & 2.36 & 12.92 & 93.76 \\
\textbf{gpt-4.1-nano+fs} & 75.39 & 11.18 & 3.20 & 4.66 & 93.76 \\
\textbf{gpt-4.1-mini} & 92.42 & 5.28 & 1.18 & 1.12 & \textbf{99.78} \\
\textbf{gpt-4.1-mini+fs} & 92.64 & 5.28 & 1.24 & \textbf{0.84} & \textbf{99.78} \\
\textbf{deepseek-chat-v3-0324} & 94.04 & 3.65 & 1.29 & 1.01 & 99.78 \\
\textbf{deepseek-chat-v3-0324+fs} & \textbf{94.21} & 3.65 & 1.29 & \textbf{0.84} & \textbf{99.78} \\
\textbf{deepseek-r1} & 91.80 & 5.11 & 1.69 & 1.40 & \textbf{99.78} \\
\textbf{deepseek-r1+fs} & 91.97 & 5.11 & 1.69 & 1.24 & \textbf{99.78} \\
\textbf{qwen-2.5-72b-instruct} & 89.61 & 6.40 & 1.01 & 2.25 & 98.99 \\
\textbf{qwen-2.5-72b-instruct+fs} & 90.28 & 6.52 & 1.07 & 1.40 & 98.99 \\
\bottomrule
\multicolumn{6}{@{}p{\textwidth}@{}}{\footnotesize \textit{Note}: Bold values indicate the highest scores in the Exact Match column, the lowest values in the Mismatched column, and the highest percentages in the Spans Checked column.} \\
\end{tabularx}
\end{table}

 Lastly, we would like to compare the transferring task in our first view and those we ended up with. The~first one that implies using indices consists of the next nonexhaustive cognitive steps: in the English text, find the symbols according to the start index and the end symbol, select symbols in between, translate the resulting substring into another language, locate in the text in another language the corresponding text, and~determine the indices in such a way that the final substring will be necessary and sufficient. On~the other hand, in~the variant where only substrings are used, the~steps are similar except there are no steps with the index-related operations, so we can hypothesize that the second task is easier from a cognitive perspective. The LLM fails on the first task, given that it is not bad at math~\cite{Dao2023InvestigatingTE}, so including counting objects, we might say that the task complexity is accounted for, as the number of cognitive steps is crucial for the LLM to complete the task~successfully. % EE: Please check that the intended meaning has been retained. %AUTHORS: confirmed

\subsection{Text~Translation}
The results for the two prompts that we used are shown in Table~\ref{translation_result}. We tested them only on gpt-3.5-turbo and llama-3.1-8b at the earlier stage of the research.  It is clear that prompts perform very closely in terms of all metrics. Regarding the BLEU score for gpt-3.5-turbo, results for this task are notably higher than the result obtained in work~\cite{Zhu2023MultilingualMT} where the ChatGPT obtains a BLEU score of around 45 points on the Eng--Rus pair. We also manually checked several dozen random examples to ensure the sanity of the translation. The~text of the two prompts can be found in Appendixes \ref{app_translation_prompt_1} and  \ref{app_translation_prompt_2}.
At the late stage of the work, when we start experimenting with other LLMs, we use prompt 2, as it shows the best performance on gpt-3.5-turbo. We are aware that one prompt might show different results depending on the LLM, but our results in Table~\ref{translation_result} show that this difference is small despite the difference in LLM scale. As~a result, after~comparing other LLMs, gpt-4o-mini and deepseek-chat-v3-0324 demonstrate the best performance, as~shown in Table~\ref{translation_result_prompt2}.

% First table of prompts comparison on text translation
\begin{table}[H]
\centering
\caption{The %MDPI: We completed the mid-lines as required. Please confirm. Same for following tables. %AUTHORS: Thank you, we confirm.
 results of the text~translation.}
\label{translation_result}
\begin{tabularx}{\textwidth}{@{}l*{6}{C}@{}}
\toprule
\textbf{} & \multicolumn{4}{c}{\textbf{LaBSE}} & \multicolumn{2}{c@{}}{\textbf{BLEU}} \\
 \cmidrule{2-7}
 & \multicolumn{2}{c}{\textbf{BLEU-like}} & \multicolumn{2}{c}{\textbf{Parallel Comparision}} & \multicolumn{2}{c@{}}{\textbf{BLEU Score}} \\
 \cmidrule{2-7}

\textbf{} & \textbf{gpt-3.5-turbo} & \textbf{llama-3.1-8b} & \textbf{gpt-3.5-turbo} & \textbf{llama-3.1-8b} & \textbf{gpt-3.5-turbo} & \textbf{llama-3.1-8b} \\
\midrule
Prompt 1 & 0.2267 & 0.2806 & 0.0010 & $-$0.0069 & 0.5011 & 0.4076 \\
Prompt 2 & 0.2288 & 0.2834 & 0.0005 & $-$0.0090 & 0.4993 & 0.4051 \\
\bottomrule
\end{tabularx}
\end{table}
\unskip

% Добавил таблицу по всем моделям с замерами на GOLD (only Prompt 2)
\begin{table}[H]
\centering
\caption{The %MDPI: Please add an explanation for the use of bold in the table footer. If the bold is unnecessary, please remove it. %AUTHORS: We added the explanation about using bold values in the table.
 results of the text translation for prompt~2.}
\label{translation_result_prompt2}
\begin{tabularx}{\textwidth}{@{}l*{2}{C}C@{}}
\toprule
 & \multicolumn{2}{c}{\textbf{LaBSE}} & \textbf{BLEU} \\
\cmidrule{2-4}
\textbf{Model} & \textbf{BLEU-like} & \textbf{Parallel Comparison} & \textbf{BLEU Score} \\
\midrule
\textbf{llama-3.1-8b}          & 0.2834          & $-$0.0090          & 0.4051 \\
\textbf{gpt-3.5-turbo}         & 0.2288          & \textbf{0.0005}  & 0.4993 \\
\textbf{gpt-4o-mini}           & \textbf{0.2168} & \textbf{0.0011}  & \textbf{0.5277} \\
\textbf{gpt-4.1-nano}          & 0.2383          & $-$0.0071          & 0.4650 \\
\textbf{gpt-4.1-mini}          & 0.2227          & $-$0.0039          & 0.4971 \\
\textbf{deepseek-chat-v3-0324} & \textbf{0.2140} & \textbf{$-$0.0019} & \textbf{0.5254} \\
\textbf{deepseek-r1}           & 0.2371          & $-$0.0081          & 0.4625 \\
\textbf{qwen-2.5-72b-instruct} & 0.2468          & $-$0.0066          & 0.4567 \\
\bottomrule
\multicolumn{4}{@{}p{\textwidth}@{}}{\footnotesize \textit{Note}: Bold values indicate the best results in each column (highest scores for BLEU-like and BLEU Score, most favorable values for Parallel Comparison), including cases with multiple similar top values.} \\  
\end{tabularx}
\end{table}
\unskip

\subsection{Whole Transferring and Model~Training}
For the complete dataset transfer, we chose the deepseek-chat-v3-0324 model because, while it does not show the best performance in translation, with~a little gap from the best-performing model, it shows the best performance in NER annotation transferring, which is the key operation. The~time spent on applying the pipeline to the whole DEFT corpus and WCL dataset is presented in Table~\ref{deepseek_chat_v3_cost_of_dataset}. On~the output, we obtain the translated corpus with NER annotation. The~statistics of the dataset for two tasks are presented in Figure~\ref{deft_stat} in the ''RUS'' legend parts.
Next, we use this dataset to train the BERT-like models to establish the baseline for the task of definition detection (Task 1) and named entity recognition (Task 2) for definition and term span detection (for the WCL dataset, the only term span detection). Our base model list is~next:
\begin{itemize}
    \item BERT-base-multilingual~\cite{DBLP:journals/corr/abs-1810-04805}---the BERT model trained by Google. A~good baseline.
    \item RuBERT-base-cased~\cite{Kuratov2019AdaptationOD}---RuBERT pretrained from scratch on Russian texts.
    \item \textls[-15]{RoBerta-base~(\url{https://huggingface.co/blinoff/roberta-base-russian-v0}, accessed on 24 April 2025)---RoBERTa~\cite{Liu2019RoBERTaAR}} %MDPI: 1. Footnotes are not supported in our journal. We have therefore included this paragraph in the main text. Please confirm. 2. Please provide the date you accessed the URL in the following format: “URL (accessed on Day Month Year)”. %AUTHORS: 1. Thank you, we confirm. 2. We provided the date of access to the URL.
 model pretrained on Russian texts.
\end{itemize}

The results of training on the RuDEFT dataset are presented in Table~\ref{model_result_rudeft} and training on WCL-Wiki-Ru in Table~\ref{model_result_wcl_wiki_ru}. For~the RuDEFT, we can see that the models achieve quite good results in the detection of sentences with definitions. For~Task 2, the~results are notably weaker, which implies that one may need to verify the correctness of the annotation to improve the recognition quality. It is interesting that RoBERTa shows such bad performance. We also see that the difference between the multilingual BERT and RuBERT is~insignificant. 

Comparing the manually revisited gold part and LLM-translated test parts, we can see a notable gap between them in definition detection. It suggests that while translating, the LLM induces some sort of bias in the text, which the classification model exploits during~training.

On the other hand, there is an interesting behavior in Task 2. First, the~gap is much lower in general across all models. Second, the~models show better results on the gold part. Third, while the gap between RuBERT and Bert-m is stable for Task 1, for~Task 2, Bert-m shows a lower gap than RuBERT. If~we take a look at the confusion matrix in Appendix~\ref{confusion_matrices}, we will see that the difference primarily comes from the better recognition of the I-Term in the gold part. Probably, this is related to the fact that NER transfer contains partially wrong spans (narrower or wider), but~the models managed to generalize in the right way on the transferred train part, although~it also contains erroneous spans. The~difference between Bert-m and RuBERT might be explained by the fact that RuBERT, which is completely trained on Russian data, can better recognize the nuances of the language, which makes it more robust to the annotation~errors. 

We leave a detailed analysis of these two phenomena for future~work. 

As for the WCL-Wiki-Ru dataset, we can see in Table~\ref{model_result_wcl_wiki_ru} that all models show excellent results on both tasks. Note that RoBERTa is slightly worse than other models. Such a good result might be explained by the simplicity of the dataset itself. As~the definitions come from Wikipedia, they have a well-defined structure. If~one looks at the definitions, one notices that they are structured like ``The X is/named/etc.'', that is, the term is placed at the beginning of the sentence, which is followed by the verb. Also, the~dataset contains only a Term entity, which makes the whole task~easier.

% Old table before review
% \begin{table}[H]
% \centering
% \caption{The results of the model training. Task 1 - definition detection, Task 2 - Term and Definition recognition.}
% \label{model_result}
% \begin{tabularx}{\textwidth}{@{}l*{6}{C}@{}}
% \toprule
% \textbf{Metrics} & \multicolumn{3}{c}{\textbf{Task 1}} & \multicolumn{3}{c}{\textbf{Task 2}} \\
% \cmidrule(lr){2-4} \cmidrule(lr){5-7}
%  & rubert & roberta & bert-m & rubert & roberta & bert-m \\
% \midrule
% Precision & 0.82 & 0.73 & 0.81 & 0.58 & 0.52 & 0.61 \\
% Recall    & 0.86 & 0.82 & 0.87 & 0.46 & 0.33 & 0.43 \\
% F1        & 0.84 & 0.76 & 0.83 & 0.51 & 0.40 & 0.51 \\
% \bottomrule
% \end{tabularx}
% \end{table}

% New table model training rudeft
\begin{table}[H]
\centering
\caption{The results of the model training on RuDEFT. Task 1---definition detection, Task 2---term and definition~recognition.}
\label{model_result_rudeft}
\begin{tabularx}{\textwidth}{@{}l *{6}{C} @{}}
\toprule
 & \multicolumn{2}{c}{\textbf{Rubert}} 
 & \multicolumn{2}{c}{\textbf{Roberta}} 
 & \multicolumn{2}{c}{\textbf{Bert-m}} \\
\cmidrule{2-7}
 \textbf{Metrics} %MDPI: Please check whether we need to merge the cells. %AUTHORS: There is no need to merge the cells, thank you.
 & \textbf{Gold} & \textbf{Test} & \textbf{Gold} & \textbf{Test} & \textbf{Gold} & \textbf{Test} \\
\midrule
 & \multicolumn{6}{c}{\textbf{Task 1}} %MDPI: Please add an explanation for the use of bold in the table footer. If the bold is unnecessary, please remove it. %AUTHORS: We added the explanation about using bold values in the table. The phrases "Task 1" and "Task 2" are bolded to provide enhanced visual distinction between the two evaluation tasks. However, if the editorial team believes that removing the bold formatting for these headers would improve readability, we kindly request to adjust it accordingly.  
 \\
% \midrule
\cmidrule(l){2-7}
Precision & 0.72 & 0.85 & 0.65 & 0.83 & 0.72 & 0.85 \\
Recall    & 0.81 & 0.85 & 0.70 & 0.76 & 0.81 & 0.83 \\
F1        & \textbf{0.73} & \textbf{0.85} & 0.57 & 0.78 & 0.72 & 0.84 \\
\midrule
 & \multicolumn{6}{c}{\textbf{Task 2}} \\
% \midrule
\cmidrule(l){2-7}
Precision & 0.73 & 0.63 & 0.68 & 0.64 & 0.71 & 0.66 \\
Recall    & 0.58 & 0.55 & 0.29 & 0.30 & 0.52 & 0.51 \\
F1        & \textbf{0.64} & \textbf{0.59} & 0.41 & 0.41 & 0.60 & 0.58 \\
\bottomrule
\multicolumn{7}{@{}p{\textwidth}@{}}{\footnotesize \textit{Note}: Bold values highlight the maximum F1 scores for each task (Task 1 and Task 2) across all models.} \\  %AUTHORS: We added the explanation about using bold values in the table.
\end{tabularx}
\end{table}
\unskip

% New table model training wcl_wiki_ru (only test set)
\begin{table}[H]
\centering
\caption{The results of the model training on WCL-Wiki-Ru. Task 1---definition detection, Task 2---term and definition~recognition.}
\label{model_result_wcl_wiki_ru}
\begin{tabularx}{\textwidth}{@{}X *{6}{C}@{}}
\toprule
 & \multicolumn{3}{c}{\textbf{Task 1}} & \multicolumn{3}{c}{\textbf{Task 2}} \\
\cmidrule{2-7}
\textbf{Metrics}%MDPI: Please check whether we need to merge the cells. %AUTHORS: There is no need to merge the cells, thank you.
 & \textbf{Rubert} & \textbf{Roberta} & \textbf{Bert-m} & \textbf{Rubert} & \textbf{Roberta} & \textbf{Bert-m} \\
\midrule
Precision & 0.96 & 0.94 & 0.96 & 0.85 & 0.87 & 0.86 \\
Recall    & 0.97 & 0.94 & 0.96 & 0.93 & 0.85 & 0.91 \\
F1        & \textbf{0.96} & 0.94 & \textbf{0.96} & \textbf{0.89} & 0.86 & \textbf{0.89} \\
\bottomrule
\multicolumn{7}{@{}p{\textwidth}@{}}{\footnotesize \textit{Note}: Bold values highlight the maximum F1 scores for each task (Task 1 and Task 2) across all models.} \\  
\end{tabularx}
\end{table}
\unskip

%%%%%%%%%%%%%%%%%%%%%%%%%%%%%%%%%%%%%%%%%%
\section{Discussion}
In this work, we show that the current abilities of LLMs can be used to transfer datasets between languages. While the dataset certainly will remain only “silver”-grade quality, the~effort difference between creating such a dataset from scratch and adapting from another language with further verification is huge. Especially when we talk about nontrivial annotations like NER that require the exact positioning in text. The~focus of our experiment was the DEFT dataset, which contains a quite challenging task of term and definition recognition. This dataset is important to facilitate the trend analysis tools and models for Russian, which are helpful for the decision-making processes in R\&D. We train the BERT-based models on the whole transferred dataset to show that these data actually can be used to train real models to establish a baseline. However, we see that the NER model is weak, which implies verifying the transferred annotation more carefully. In~addition, we apply our pipeline to the Wikipedia part of the WCL dataset and show that models show quite good~results.

As a side effect, we discover that some tasks might be easier for LLMs to understand than others, and~that difference may be significant in terms of output quality. We hypothesize that it depends on several cognitive steps that need to be performed for task solving. We also discovered some shortcomings in the DEFT dataset that should be~fixed.

The obvious limitations of our work are that we do not show how the LLM itself would be strong on DEFT tasks. Another one is that we do not compare the quality of supervised translators with LLMs, because it is shown that supervised translators are still better than~LLMs. 

%%%%%%%%%%%%%%%%%%%%%%%%%%%%%%%%%%%%%%%%%%

%%%%%%%%%%%%%%%%%%%%%%%%%%%%%%%%%%%%%%%%%%
\vspace{6pt} 

%%%%%%%%%%%%%%%%%%%%%%%%%%%%%%%%%%%%%%%%%%
%% optional
%\supplementary{The following supporting information can be downloaded at:  \linksupplementary{s1}, Figure S1: title; Table S1: title; Video S1: title.}


% Only used for preprtints:
% \supplementary{The following supporting information can be downloaded at the website of this paper posted on \href{https://www.preprints.org/}{Preprints.org}.}

%\begin{tabularx}{\textwidth}{lll}
%\toprule
%\textbf{Name} & \textbf{Type} & \textbf{Description} \\
%\midrule
%S1 & Python script (.py) & Script of python source code used in XX \\
%S2 & Text (.txt) & Script of modelling code used to make Figure X \\
%S3 & Text (.txt) & Raw data from experiment X \\
%S4 & Video (.mp4) & Video demonstrating the hardware in use \\
%\bottomrule
%\end{tabularx}
%}

%%%%%%%%%%%%%%%%%%%%%%%%%%%%%%%%%%%%%%%%%%
% D.P., E.T., D.S., I.S., I.B.
\authorcontributions{Conceptualization, I.S. and I.B.; methodology, D.P., E.T., and I.B.; software, D.P., E.T., and I.B.; validation, D.S.; formal analysis, D.P., E.T., and I.B.; investigation, D.P., E.T., and D.S.; data curation, D.P. and E.T.; writing---original draft preparation, D.P., E.T., and D.S.; writing---review and editing, I.S. and I.B.; visualization, D.P., E.T., and D.S.; supervision, I.S. and I.B.; project administration, I.S. and I.B. All authors have read and agreed to the published version of the manuscript.}

% \funding{This research received no external~funding.}

\funding{The experimental study was supported by the Project of the Research Center for Trusted Artificial Intelligence of the Ivannikov Institute for System Programming of the Russian Academy of Sciences.} %AUTHORS: The funding information has been changed.

\dataavailability{The data generated in this study are available at Hugging Face: %MDPI: Please provide the date you accessed the URL in the following format: “URL (accessed on Day Month Year)”. %AUTHORS: We provided the date of access to the URL.
 \url{https://huggingface.co/datasets/astromis/ruDEFT} (accessed on 24 April 2025), \url{https://huggingface.co/datasets/astromis/WCL_Wiki_Ru} (accessed on 24 April 2025). The~source code used in this work is publicly available on GitHub: \url{https://github.com/Astromis/research/tree/master/rudeft} (accessed on 24 April 2025) . These links ensure open access to the data and code in accordance with MDPI's research data policies. The~full policy is available at \url{https://www.mdpi.com/ethics} (accessed on 24 April 2025).}

\acknowledgments{We would like to thank the reviewers for their valuable comments that helped us to improve our paper.} %AUTHORS: The Acknowledgments section has been changed.

\conflictsofinterest{The authors declare no conflicts of~interest.} 

%%%%%%%%%%%%%%%%%%%%%%%%%%%%%%%%%%%%%%%%%%
%% Optional

%% Only for journal Encyclopedia
%\entrylink{The Link to this entry published on the encyclopedia platform.}

% \abbreviations{Abbreviations}{
% The following abbreviations are used in this manuscript:\\

% \noindent 
% \begin{tabular}{@{}ll}
% MDPI & Multidisciplinary Digital Publishing Institute\\
% DOAJ & Directory of open access journals\\
% TLA & Three letter acronym\\
% LD & Linear dichroism
% \end{tabular}
% }

%%%%%%%%%%%%%%%%%%%%%%%%%%%%%%%%%%%%%%%%%%
%% Optional
\appendixtitles{yes} % Leave argument "no" if all appendix headings stay EMPTY (then no dot is printed after "Appendix A"). If~the appendix sections contain a heading then change the argument to "yes".
\appendixstart
\appendix
\section{Prompt for Annotation~Transferring}
\label{app_ner_transferring_prompt}

Given a JSON object, find the exact corresponding text in the Russian translation for each English span and store the results in a new field called spans\_rus. %MDPI:  it's suggested to unify all the font to be Palatino Linotype. We highlighted only a few cases as a reminder, please check the whole text carefully. %AUTHORS: We replaced all cases to the font Palatino Linotype.
 The~input JSON object contains the following~fields:
\begin{itemize}
    \item text:  English source text.
    \item text\_rus: Russian translated text.
    \item spans: A list of spans, each containing: 1. The~start index in the English text; 2.~The~end index in the English text; 3. The~label; 4. The~ID; 5. The~portion of the English text that was extracted using the start and end indices.
    \item And other fields.
\end{itemize}

Your task is to: For each span, locate the exact corresponding Russian text %MDPI: Please confirm if the bold formatting is necessary; if not, please remove it. The following highlights are the same. %AUTHORS: We removed the bold formatting.
 in text\_rus that matches the exact wording of the English span (the 5th element in each span) in~meaning.

Important:
\begin{itemize}
    \item Do not modify or correct the form, word order, or~any grammatical aspects of the Russian text — it must be extracted exactly as it appears in text\_rus, including word endings, grammatical cases, punctuation, punctuation marks, and spacing.
    \item Record the matched Russian text as a new list in a new field spans\_rus, where each item is also a list containing the same label and ID as in the English span, and~the matched Russian text.
    \item For each span in spans, one must obtain a span in spans\_rus.
\end{itemize}

No %MDPI: We cahnged the indent as required. please confirm. %AUTHORS: Thank you, we confirm.
 explanation, just output the updated~JSON.

\section{Prompt 1 for~Translation}
\label{app_translation_prompt_1}

Given a JSON object, write an accurate translation into Russian for the original English sentence and save the results in a new field named text\_rus. The~input JSON object contains the following~fields:
\begin{itemize}
    \item id: Unique ID of sentence.
    \item text: English source text.
\end{itemize}

Your task is to: For each text in English (text) write its exact translation into Russian in a scientific lexical style and save the results in a new field named text\_rus.

Important:
\begin{itemize}
    \item Write down the corresponding translated Russian text in the form of a new text\_rus field.
    \item For English text in text, one should definitely obtain the Russian text in text\_rus.
\end{itemize}

No %MDPI: We cahnged the indent as required. please confirm. %AUTHORS: Thank you, we confirm.
 explanation, just output the updated~JSON.

\section{Prompt 2 for~Translation}
\label{app_translation_prompt_2}

Given a JSON object, write an accurate translation into Russian for the original English sentence and save the results in a new field named text\_rus. The~input JSON object contains the following~fields:
\begin{itemize}
    \item id: Unique ID of sentence.
    \item text: English source text.
\end{itemize}

Your task is to: For each text in English write its exact translation into Russian taking into account the style of the sentence and its scientific significance (for example, medical, historical, etc.) and save the results in a new field named text\_rus.

Important:
\begin{itemize}
    \item Write down the corresponding translated Russian text in the form of a new text\_rus field.
    \item For English text in text, one should definitely obtain the Russian text in text\_rus.
\end{itemize}

No explanation, just output the updated~JSON. %AUTHORS: We changed the indent as required.

\section{Confusion~Matrices}
\label{confusion_matrices}

\vspace{-10pt}
\begin{figure}[H]
\includegraphics[width=\textwidth,height=\textheight,keepaspectratio]{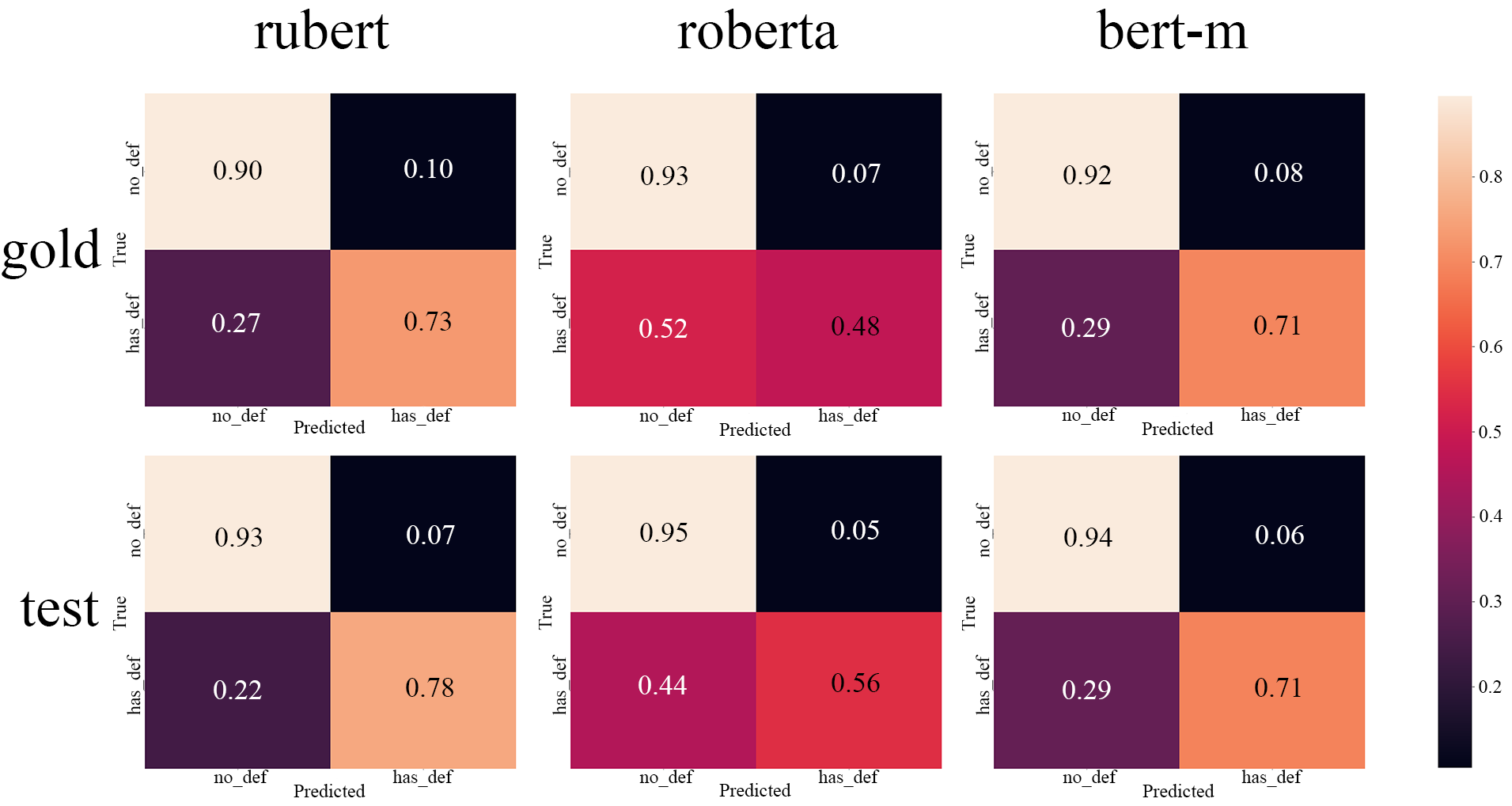}
\caption{Confusion matrices for Task 1 on the RuDEFT~dataset.}\label{heatmap_rudeft_task}
\end{figure}
\unskip

\begin{figure}[H]

\includegraphics[width=\textwidth,height=\textheight,keepaspectratio]{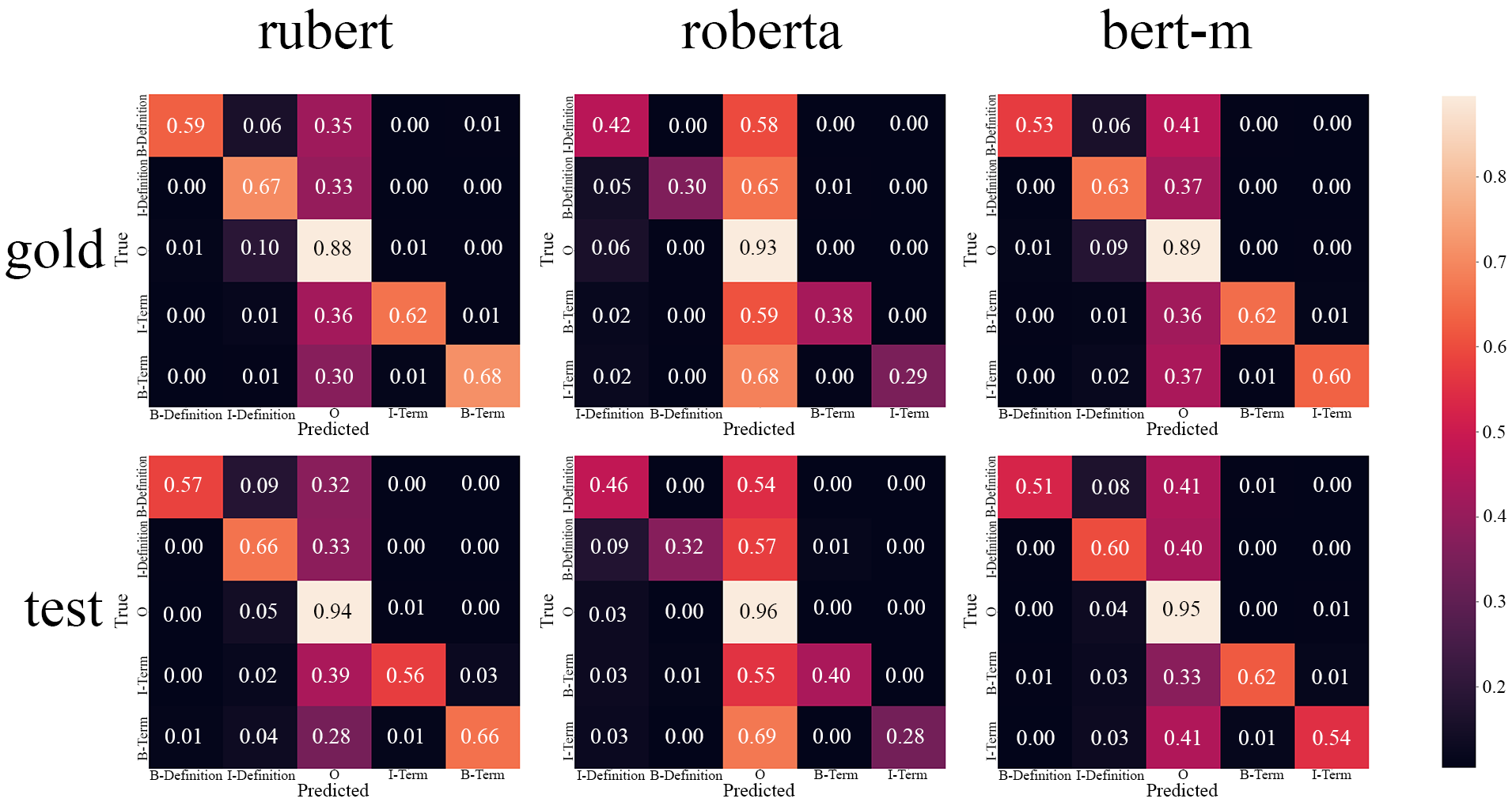}
\caption{Confusion matrices for Task 2 on the RuDEFT~dataset.}\label{heatmap_rudeft_task2}
\end{figure}
\unskip

\begin{figure}[H]

\includegraphics[width=\textwidth,height=\textheight,keepaspectratio]{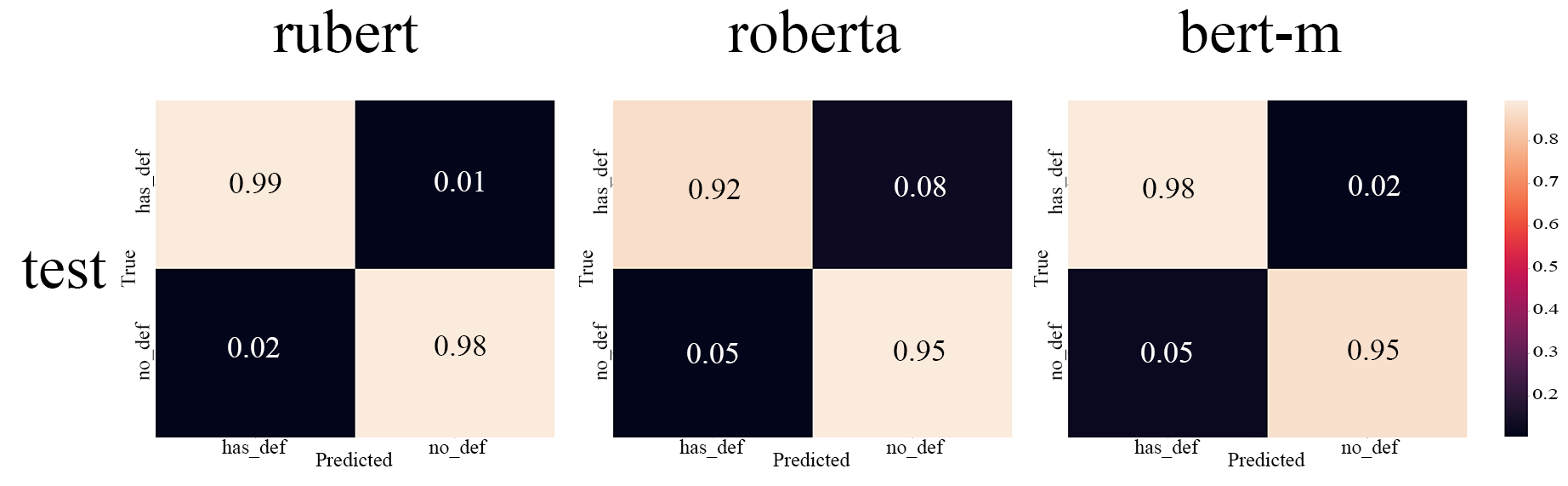}
\caption{Confusion matrices for Task 1 on the Wikipedia part of the WCL~dataset.}\label{heatmap_wiki_task1}
\end{figure}
\unskip

\begin{figure}[H]

\includegraphics[width=\textwidth,height=\textheight,keepaspectratio]{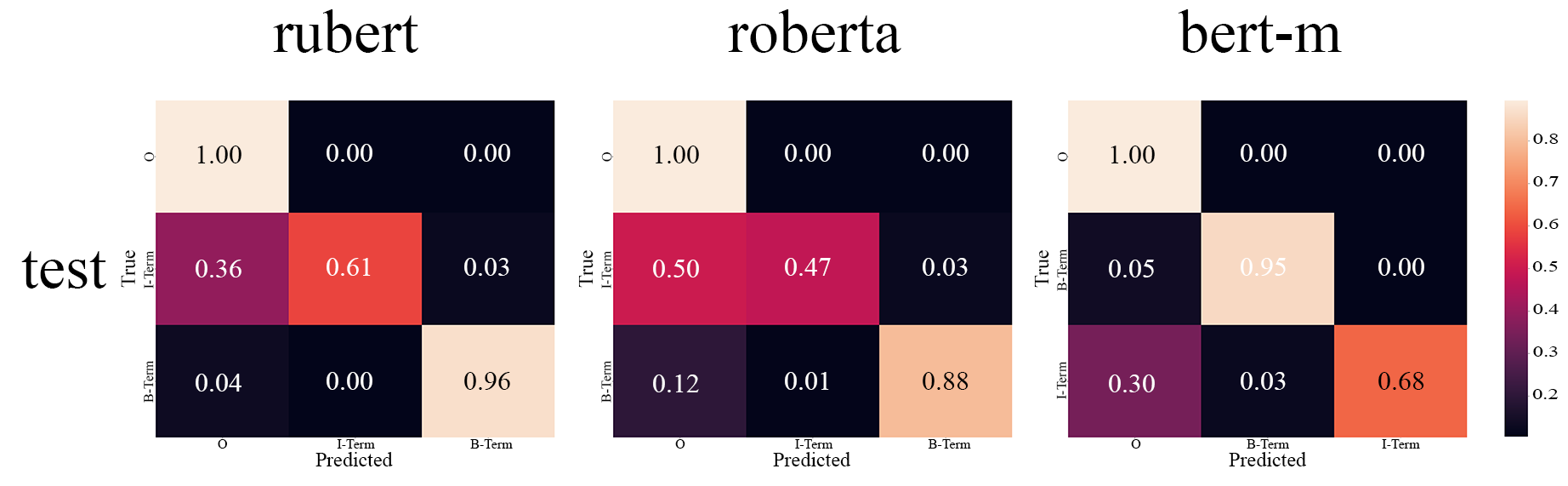}
\caption{Confusion matrices for Task 2 on the Wikipedia part of the WCL~dataset.}\label{heatmap_wiki_task2}
\end{figure}

\section{LLM Usage: Time and Cost Statistics Across~Tasks}
\label{appendix:costs}

\vspace{-6pt}
% Table of Comparison of model metrics (avg cost, avg time)
\begin{table}[H]
\centering
\caption{Comparison %MDPI: 1. Please confirm if the bold formatting of first column is necessary; if not, please remove it. The same applies below as well to bold in table. 2. We completed mid-line as required. Please confirm. %AUTHORS: 1. The bold formatting in the first column is used to enhance visual distinction of row headers. However, if the editorial team recommends removing the bold for consistency, we defer to your judgment. 2. Thank you, we confirm.
 of model metrics. CAPS/ex.---CAPS per example (CAPS---the internal currency of the bothub.chat service), USD/ex.---Dollars per example, Time/ex.---Time per example in~seconds.}
\label{model_cost_comparison}
\begin{tabularx}{\textwidth}{@{}l *{6}{C}@{}}
\toprule
 & \multicolumn{3}{c}{\textbf{Text Translation}} & \multicolumn{3}{c}{\textbf{NER Transferring}} \\
\cmidrule{2-7}
\textbf{Model} %MDPI: Please check whether we need to merge the cells. %AUTHORS: There is no need to merge the cells, thank you.
 & \textbf{CAPS/ex.} & \textbf{USD/ex.} & \textbf{Time/ex. (s) %MDPI: We changed sec into s which refers to second as required. Please confirm. %AUTHORS: Thank you, we confirm.
} & \textbf{CAPS/ex.} & \textbf{USD/ex.} & \textbf{Time/ex. (s)} \\
\midrule
\textbf{llama-3.1-8b-instruct}    & 14   & 0.00002 & 2.94    & 121   & 0.00021 & 16.8 \\
\textbf{gpt-3.5-turbo}            & 410  & 0.00072 & 1.28    & 1967  & 0.00347 & 9.6 \\
\textbf{gpt-4o-mini}              & 119  & 0.00021 & 0.03    & 627   & 0.00111 & 9.6 \\
\textbf{gpt-4.1-nano}             & 79   & 0.00014 & 0.02    & 410   & 0.00072 & 6  \\
\textbf{gpt-4.1-mini}             & 312  & 0.00055 & 0.02    & 1595  & 0.00282 & 7.2 \\
\textbf{deepseek-chat-v3-0324}    & 222  & 0.00039 & 0.14    & 1159  & 0.00205 & 36  \\
\textbf{deepseek-r1}              & 1590 & 0.00281 & 36.52   & 5132  & 0.00906 & 68.4 \\
\textbf{qwen-2.5-72b-instruct}    & 111  & 0.00020 & 0.05    & 582   & 0.00103 & 28.8 \\
\bottomrule
\end{tabularx}
\end{table}
\unskip

% Table of Resources spent on pipeline processing of datasets. (all cost, all time)
\begin{table}[H]
\centering
\caption{Resources spent on pipeline processing of~datasets.}
\label{deepseek_chat_v3_cost_of_dataset}
\begin{tabularx}{\textwidth}{@{}l *{6}{C} @{}}
\toprule
 & \multicolumn{3}{c}{\textbf{Text Translation}} & \multicolumn{3}{c}{\textbf{NER Transferring}} \\
\cmidrule{2-7}
 \textbf{Dataset } %MDPI: Please check whether we need to merge the cells. %AUTHORS: There is no need to merge the cells, thank you.
& \textbf{CAPS} & \textbf{Dollars (USD)} & \textbf{Time (min)} & \textbf{CAPS} & \textbf{Dollars (USD)} & \textbf{Time (min)} \\
\midrule
\textbf{RuDEFT}        & 6,497,571 %MDPI: We added commas to separate out the thousands for numbers with five or more digits. Please confirm. %AUTHORS: Thank you, we confirm.
 & 11.5 & 503 &13,125,904 & 23.2 & 608 \\
\textbf{WCL-Wiki-Ru}   & 1,030,348 & 1.8  & 49  & 2,031,111  & 3.6  & 98  \\
\bottomrule
\end{tabularx}
\end{table}

\section{Software and~Dependencies}
\label{appendix:software}

All experiments and data processing were carried out using Python 3.11. %MDPI: Please confirm if the bold formatting is necessary; if not, please remove it. %AUTHORS: We removed the bold formatting.
  The~following libraries, frameworks, and~external APIs were installed with the specified versions to ensure full reproducibility:

\medskip
\begin{itemize}
  \item langchain==0.3.1
  \item pydantic==2.9.2
  \item fuzzysearch==0.7.3
  \item datasets==2.21.0 (Hugging Face)
  \item sentence\_transformers==3.1.1
  \item scipy==1.14.0
  \item evaluate==0.4.3
  \item label-studio==1.12.1
\end{itemize}

For %MDPI:  We cahnged the indent as required. please confirm. %AUTHORS: Thank you, we confirm.
 a complete list of all dependencies and their exact versions, please refer to the requirements.txt file in the project repository on \href{https://github.com/Astromis/research/tree/master/rudeft}{GitHub}.

In %MDPI: We cahnged the indent as required. please confirm. %AUTHORS: Thank you, we confirm.
 addition, we leveraged external services and APIs for specific~tasks:
\begin{itemize}
  \item Google Translate API %MDPI: Please confirm if the italics are necessary; if not, please remove them. The following highlights are the same. %AUTHORS: We removed the italics.
 for automated text translation.
  \item bothub.chat API as a unified proxy for accessing multiple model families.
\end{itemize}

%%%%%%%%%%%%%%%%%%%%%%%%%%%%%%%%%%%%%%%%%%
%\isPreprints{}{% This command is only used for ``preprints''. 
\begin{adjustwidth}{-\extralength}{0cm}

\reftitle{References} %MDPI: Newly added and/or altered information is highlighted. Please confirm. %AUTHORS: newly information confirmed or changed. Thank you.

\PublishersNote{}
%\isPreprints{}{% This command is only used for ``preprints''.
\end{adjustwidth}
\end{document}